\documentclass[conference]{IEEEtran}
\IEEEoverridecommandlockouts
\usepackage{cite}
\usepackage{amsmath,amssymb,amsfonts}
\usepackage{algorithmic}
\usepackage{graphicx}
\usepackage{textcomp}
\usepackage{xcolor}
\usepackage{cite}
\usepackage{amsmath,amssymb,amsfonts}
\usepackage{algorithmic}
\usepackage{graphicx}
\usepackage{caption}
\usepackage{textcomp}
\usepackage{xcolor}
\usepackage{svg}
\usepackage{afterpage}
\usepackage{booktabs}
\usepackage{float}
\usepackage{subfigure}
\usepackage{multirow}
\usepackage{todonotes}
\usepackage{soul}
\usepackage{lipsum}
\usepackage{graphicx}
\renewcommand{\baselinestretch}{0.904}
\usepackage[citecolor=blue,colorlinks=true,linkcolor=blue]{hyperref}%
\def\BibTeX{{\rm B\kern-.05em{\sc i\kern-.025em b}\kern-.08em
    T\kern-.1667em\lower.7ex\hbox{E}\kern-.125emX}}
\begin{document}

\title{FINN-GL: Generalized Mixed-Precision Extensions for FPGA-Accelerated LSTMs \\
\thanks{Vitis, Vivado, PYNQ and combinations thereof are trademarks of Advanced Micro Devices, Inc.}
}
\author{
\IEEEauthorblockN{Shashwat Khandelwal\IEEEauthorrefmark{1}, Jakoba Petri-Koenig\IEEEauthorrefmark{2}, Thomas B. Preußer\IEEEauthorrefmark{2}, Michaela Blott\IEEEauthorrefmark{2}, Shanker Shreejith\IEEEauthorrefmark{1}}
\IEEEauthorblockA{\IEEEauthorrefmark{1}Reconfigurable Computing Systems Lab, Electronic \& Electrical Engineering\\
Trinity College Dublin, Ireland, 
Email: \{khandels, shankers\}@tcd.ie}
\IEEEauthorblockA{\IEEEauthorrefmark{2}Advanced Micro Devices (AMD) Research, Dublin, Ireland\\
Email: \{Jakoba.Petri-Koenig, thomas.preusser, michaela.blott\}@amd.com}
}

\maketitle

\begin{abstract}
Recurrent neural networks (RNNs), particularly LSTMs, are effective for time-series tasks like sentiment analysis and short-term stock prediction. 
However, their computational complexity poses challenges for real-time deployment in resource constrained environments. 
While FPGAs offer a promising platform for energy-efficient AI acceleration, existing tools mainly target feed-forward networks, and LSTM acceleration typically requires full custom implementation.
In this paper, we address this gap by leveraging the open-source and extensible FINN framework to enable the generalized deployment of LSTMs on FPGAs.
Specifically, we leverage the \textit{Scan} operator from the Open Neural Network Exchange (ONNX) specification to model the recurrent nature of LSTM computations, enabling support for mixed quantisation within them and functional verification of LSTM-based models. 
Furthermore, we introduce custom transformations within the FINN compiler to map the quantised ONNX computation graph to hardware blocks from the HLS kernel library of the FINN compiler and Vitis HLS. 
We validate the proposed tool-flow by training a quantised ConvLSTM model for a mid-price stock prediction task using the widely used dataset and generating a corresponding hardware IP of the model using our flow, targeting the XCZU7EV device.
We show that the generated quantised ConvLSTM accelerator through our flow achieves a balance between performance (latency) and resource consumption, while matching (or bettering) inference accuracy of state-of-the-art models with reduced precision. 
We believe that the generalisable nature of the proposed flow will pave the way for resource-efficient RNN accelerator designs on FPGAs.
\end{abstract}

\section{Introduction}\label{sec:introduction}

\begin{figure}[t!]
    \includegraphics[scale = 0.75]{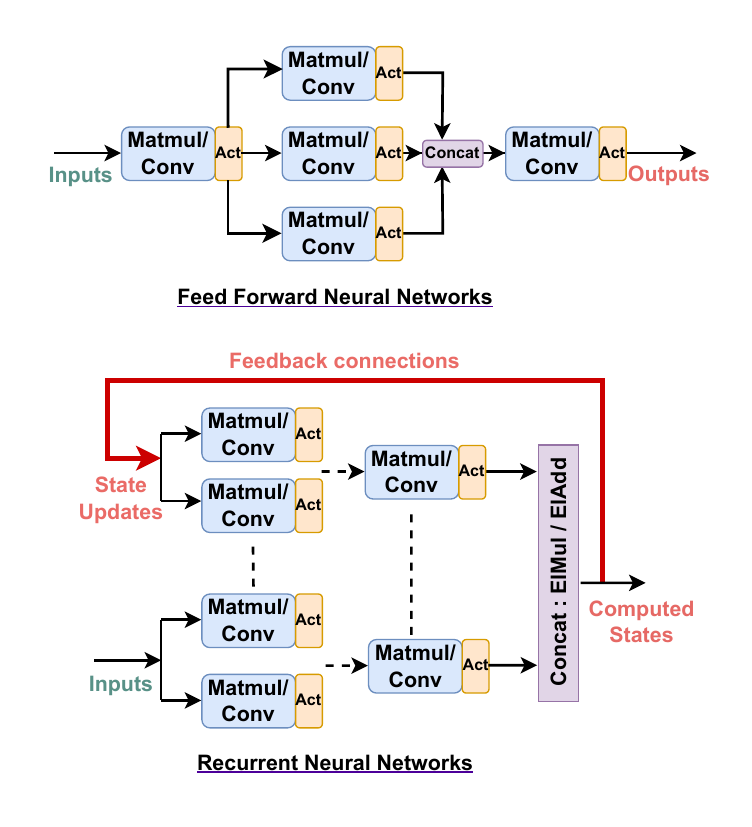}
    \caption{The figure contrasts the information flows of conventional feed-forward vs. recurrent neural networks. While in feed-forward networks, data flows acyclically from input to output, recurrent neural networks utilise the feedback of previously computed state to track sequential dependencies.}
    \label{fig:intro-fig}
\end{figure}

Time-series predictions are increasingly gaining recognition for their ability to allow stakeholders to dynamically optimise resource allocation in real-time, in applications such as energy forecasting and stock market trend prediction, among others. 
The generalisability of machine learning (ML) models, coupled with the increasing availability of high-quality training data, has led to significant improvements in prediction accuracy compared to traditional hand-tuned algorithms~\cite{zhang2019deeplob}. 
However, the real-time nature of these applications necessitates deployment on hardware accelerators to meet latency and performance requirements.
Field-programmable gate arrays (FPGAs) have emerged as a preferred choice for accelerating ML models in scenarios where energy efficiency, high throughput, and runtime reconfigurability are desirable. 
Over the past five years, several frameworks have been developed to streamline the prototyping and deployment of ML models, including convolutional neural networks (CNNs) and unconventional feed-forward architectures (such as, ResNets, Inception nets and U-Nets), by leveraging quantisation techniques to optimise their hardware performance~\cite{blott2018finn,Duarte:2018ite}.
Despite these advancements, a significant gap remains in the deployment of recurrent neural networks (RNNs), such as long short-term memory (LSTM) networks and gated recurrent units (GRUs). 

RNNs are ideally suited for time-series data processing, often outperforming feed-forward networks in extracting meaningful features and achieving higher predictive accuracy~\cite{joseph2023near}. 
Unlike feed-forward architectures, RNNs compute hidden states from a sequence of inputs using recurrent or feedback connections, making their computational graphs inherently sequential. 
Figure~\ref{fig:intro-fig} contrasts the information flows in feed-forward networks compared to RNNs. 
This capability makes them particularly well-suited for time-series applications, such as energy consumption forecasting in power grids and stock price prediction in high-frequency trading (HFT).
Such applications often demand an energy-efficient, low-latency acceleration of LSTM models on platforms such as FPGAs, where custom operators, precisions and sparsity can be exploited during the design and implementation phase. 
However, the recurrent dataflow limits the ability of existing frameworks to support efficient RNN deployments. 

Popular neural network representation formats, such as Open Neural Network Exchange (ONNX), simplify deployment by providing pre-packaged LSTM nodes that abstract away the recurrence~\cite{onnx}. 
While this abstraction facilitates standardisation and interoperability, it restricts customisation of the internal data flow, particularly for applying fine-grained quantisation or optimising specific hardware implementations. 
Consequently, (a) ONNX’s limited support for custom operations with feedback connections poses a challenge for development, and (b) the sequential nature of RNNs inherently limits parallelisation in FPGA-based implementations~\cite{rybalkin2018finn}.
Hence, existing toolchains have seen limited generalisable extensions to support models with recurrent connections.
On the contrary, developing a standalone toolchain for generalized RNN deployment on FPGAs will likely see limited adoption. 
A more practical approach is the extension of established open-source frameworks to support RNNs. 
This work aims at addressing this challenge by enhancing the RNN deployment capabilities within widely-used FPGA-based ML frameworks, paving the way for efficient, scalable, and energy-conscious solutions for real-time time-series applications.


\begin{figure*}[t!]
  \includegraphics[scale = 0.58 ]{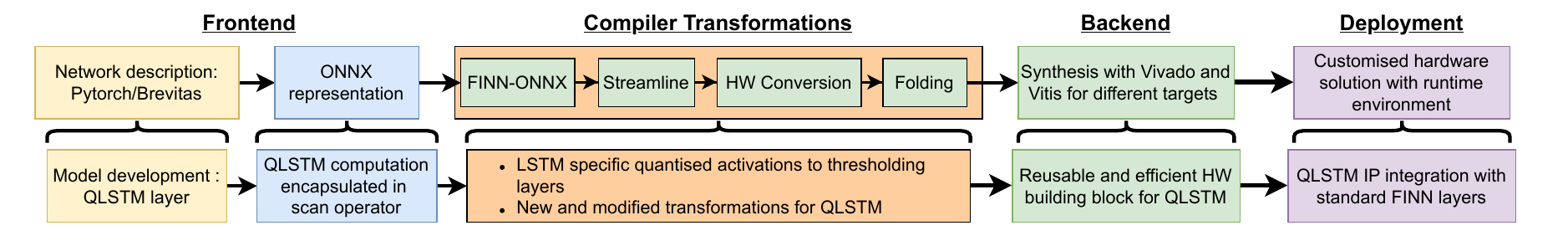}
  \caption{This figure illustrates the different stages in the  overall deployment workflow for generalized quantised LSTM layer (with mixed quantisation support) based models as an extension of the FINN framework for FPGAs. This figure is inspired by Figure 6 in~\cite{elastic-df}.}
  \label{fig:deployment-flow}
\end{figure*}

In this paper, we extend the widely used open-source FINN framework to support the generalized development and deployment of LSTM networks on FPGAs.
Figure~\ref{fig:deployment-flow} illustrates our contributions in relation with the end-to-end FINN flow. 
Through these contributions, we enable the efficient and flexible FPGA-based deployment of LSTMs, effectively addressing the challenges posed by their recurrent structure and complex computations. 
Our specific contributions in this work are outlined below: 
\begin{itemize}
    \item \textbf{ONNX Representation}:
    \begin{itemize}
        \item Development of an ONNX representation for custom LSTM models, utilising the \emph{Scan} operator to capture the recurrent nature of LSTM computations.
        The implementation is open-sourced as part of the \href{https://github.com/fastmachinelearning/qonnx/blob/main/notebooks/4_quant_lstm.ipynb}{qonnx repository}, providing a reusable and extensible framework for quantised LSTM representations.
        \item This representation integrates LSTM layers seamlessly with standard neural network layers, enabling the functional software simulation of LSTM-based networks.
        \item This representation supports mixed quantisation within an LSTM layer, broadening the scope for efficient model deployment.
    \end{itemize}
    \item \textbf{FINN Compiler Transformations}: 
    \begin{itemize}
        \item Introduction of new transformations within the FINN compiler to map LSTM computations to hardware blocks available in the HLS kernel library of the compiler (\textit{finn-hlslib}).
        \item Key transformations include: \begin{itemize}
           \item Conversion of quantised activation functions (\emph{sigmoid} and \emph{tanh}) into representations compatible with existing FINN hardware blocks, ensuring efficient hardware execution.
            \item Enhancements to network optimisation transformations (e.g., Streamlining~\cite{umuroglu2017streamlined}) to improve the hardware deployment of the LSTM compute graph, reducing resource usage and latency.
        \end{itemize}     
    \end{itemize}
     \item \textbf{HLS Layer Development}:
     \begin{itemize}
         \item Construction of hardware for LSTM layers using building blocks from \textit{finn-hlslib}, enabling a fully customisable implementation in terms of data types, dimensions, and other architectural parameters. 
         This design also enhances parallelism, improving computational efficiency while maintaining reusability for various FPGA deployments.
         \item Enabling the generation of LSTM networks as Intellectual Property (IP) blocks for FPGA designs, facilitating broader usability.
         \item The implementation of this layer is open sourced as part of the \href{https://github.com/shashwat1198/FINN-GLSTM-Hw}{FINN-GLSTM-Hw} repository.
     \end{itemize}
     \item \textbf{Deployment}:
     \begin{itemize}
         \item Demonstrating the seamless integration of an LSTM IP block within a neural network that includes convolutional and dense layers, enabling mid-price stock prediction in HFT scenarios.
     \end{itemize}
\end{itemize}

This work bridges a critical gap in the deployment of LSTMs on FPGAs, providing a comprehensive methodology for leveraging the power of RNNs in hardware-accelerated environments. 
The rest of the paper is organised as follows. Section~\ref{sec:background} discusses different LSTM implementations proposed in the literature as well as background on limit order books and the reasoning for extending FINN. Section~\ref{sec:development} discusses the development stages of the implementation. Section~\ref{sec:experiments} discusses the case study on integrating the quantised-ConvLSTM model for the mid-price stock prediction use case for HFT environments. We conclude the paper in  Section~\ref{sec:conclusion}.
\section{Background and Related Work}\label{sec:background}
\subsection{Recurrent Neural Networks}
RNNs were introduced in the 1990s primarily for sequence prediction tasks, such as speech recognition and natural language processing.
Their key advantage is the ability to maintain hidden states that capture information from previous inputs~\cite{robinson1996use}. 
This ability to model tempora    l dependencies makes them especially useful in applications where context and patterns evolve over time.
Despite their advantages, RNNs were challenging to train due to issues such as the vanishing gradient problem, which were later addressed with advanced architectures like LSTMs and GRUs. 
LSTMs and GRUs have since found widespread applications in many sequence prediction tasks. 
The compute equations (gates and states) for state computations in LSTMs are shown below:
\begin{equation}
    f_t = \sigma(W_f x_t + U_f h_{t-1} + b_f)
\end{equation}
\begin{equation}
    i_t = \sigma(W_i x_t + U_i h_{t-1} + b_i)
\end{equation}
\begin{equation}
    \tilde{C}_t = \tanh(W_c x_t + U_c h_{t-1} + b_c)
\end{equation}
\begin{equation}
    C_t = f_t \odot C_{t-1} + i_t \odot \tilde{C}_t
\end{equation}
\begin{equation}
    o_t = \sigma(W_o x_t + U_o h_{t-1} + b_o)
\end{equation}
\begin{equation}
    h_t = o_t \odot \tanh(C_t)
\end{equation}

where $\sigma$ represents the sigmoid activation function, $\tanh$ is the hyperbolic tangent activation, and $\odot$ denotes element-wise multiplication.
Due to the computationally intensive nature of these operations, LSTMs are often accelerated using bespoke hardware implementations on FPGAs to support real-time applications, as discussed in the next subsection.

\subsection{LSTMs on FPGAs}
Multiple FPGA accelerators and frameworks with RNN support have been proposed in the research literature. 
Khoda et\,al.\,\cite{khoda2023ultra} extend the \emph{hls4ml} framework by adding support for LSTM deployments on FPGAs.
They showcase three example implementations with varied LSTM layer complexity.
Que et\,al.\,\cite{que2020mapping} present a hardware architecture to address data dependency issues in LSTM computations along with a new blocking and batching strategy that enhances the weight reuse in LSTM operations. 
Their approach minimises external memory access, optimising the system performance for large LSTM-based models on devices with limited on-chip memory capacity.
F-LSTM~\cite{liang2023f} is a framework for deploying LSTM-based models on FPGAs using the CPUs to carry out pre- and post-processing. The authors test their implementation on sentiment analysis and achieve improvements over a reference GPU implementation.
Li et\,al\,\cite{10296422} propose an efficient sparse LSTM accelerator on embedded FPGAs with bandwidth-oriented pruning for reducing the on-chip memory demand.
They train and test their LSTM model on the TIMIT dataset demonstrating an optimised deployment on the PYNQ-Z1 FPGA platform.
The C-LSTM accelerator~\cite{wang2018c} leverages a structured compression technique (as opposed to random pruning) to reduce model size while avoiding the introduction of irregularities in computation and memory accesses. 
Rybalkin et\,al.\,\cite{rybalkin2018finn} propose FINN-L, an extension of FINN to deploy parametrisable quantised LSTMs on FPGAs. 
They implement a Bi-LSTM model for OCR recognition tasks on the ZCU104 FPGA platform.
This has, however, remained a custom implementation not integrated with the general FINN workflow.
Ribes et\,al.\,\cite{ribes2020mapping} propose solving the deployment constraints posed by stacked LSTM models on FPGA by altering their computational structure optimising memory requirements.
Ioannou and Fahmy~\cite{ioannou2022streaming} propose  a flexible overlay architecture for LSTMs on hybrid FPGAs. They propose a streaming dataflow arrangement exploiting the capabilities of DSP blocks while also mitigating external memory overheads.
While the above implementations introduce novel deployment architectures for LSTMs, they do not provide a generalized mixed precision development and deployment workflow, limiting the broader adoption of efficient LSTM models on FPGAs.

\subsection{FINN}
\begin{figure}[t!]
    \centering
    \includegraphics[scale = 0.45]{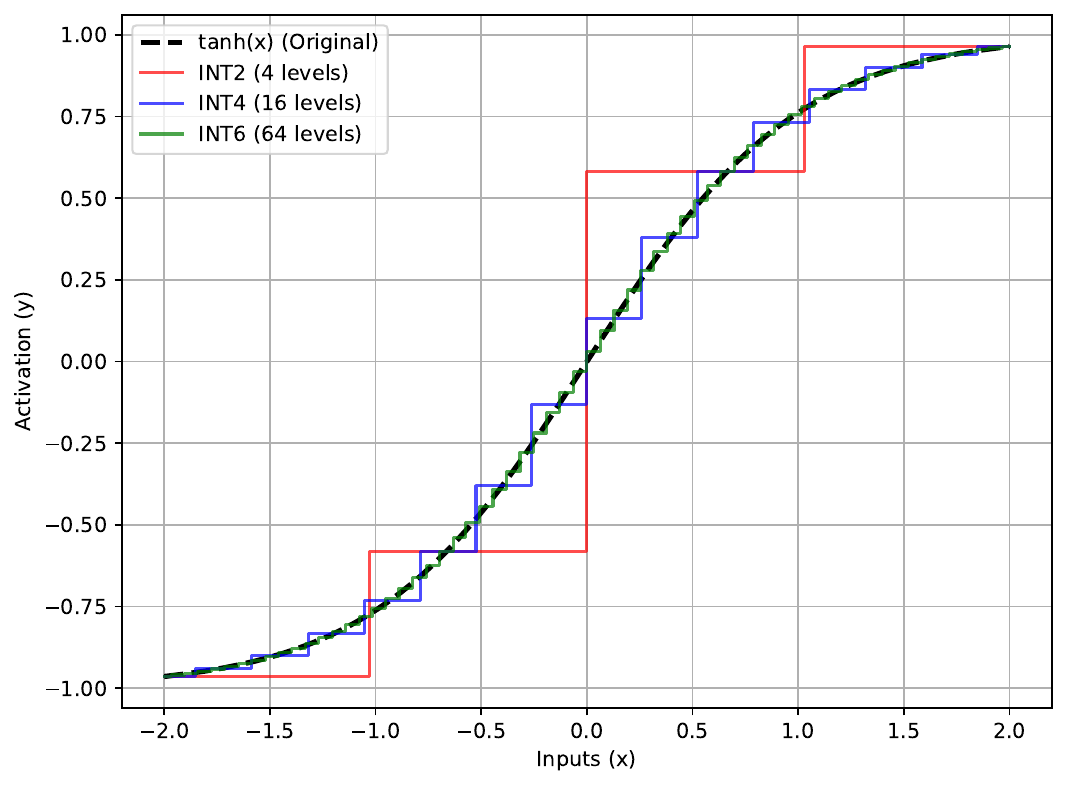}
    \caption{The figure illustrates how the Multithresholding operator approximates the tanh activation function over a defined input range for INT2, INT4 and INT6 bitwidths.}
    \label{fig:multithreshold}
\end{figure}

FINN facilitates the rapid development and deployment of quantised neural networks (QNNs) on both edge and datacenter FPGAs~\cite{blott2018finn}. 
Widely adopted by the open-source research community, it has evolved into an efficient toolchain for acceleration of quantised machine learning models. 
A distinguishing feature of this framework is its use of a \emph{thresholding} operator to implement quantised activation functions. 
This operator maps input values to integers in the interval~[0, n] by comparing the input to a set of threshold values and returning an integer value that specifies the number of threshold values that the input is greater than or equal to. 
By adjusting the spacing between threshold values, this operator can model any monotonically increasing activation function, such as, \emph{ReLU}, \emph{sigmoid}, and \emph{tanh}. 
The efficient, hardware-friendly implementation of the latter two relying on comparisons rather than the explicit computation of exponentials is key for an efficient implementation of LSTMs. 
Figure~\ref{fig:multithreshold} illustrates how to operator models the tanh activation at three different INT quantisation levels.

We demonstrate that our extensions to the FINN framework encompass all essential components, including FINN compiler transformations that map the LSTM compute graph to pre-built hardware building blocks, along with support for converting custom operators—such as mapping \emph{sigmoid} and \emph{tanh} activations to comparison based thresholding operations available through \emph{finn-hlslib}. 
These components enable the construction of hardware-accelerated LSTM layers within the framework, allowing the seamless integration into larger models.
A key advantage of quantisation using FINN is that all LSTM layer weights can be stored in the on-chip memory of the FPGA, enhancing performance and addressing limitations encountered in other implementations~\cite{10296422}.
As a result, we chose to extend this framework to develop a generalized deployment pipeline for recurrent neural networks.

\subsection{Limit Order Books}
A limit order book (LOB) is a record of pending limit orders maintained by an exchange. 
Each limit order is an instruction to buy or sell a security at a specified price or better. 
When limit orders are placed, they are recorded at the exchange by updating the order book to track them. 
Orders are executed when the market reaches or exceeds their specified price levels~\cite{lob}.
LOBs represent investors sentiments and, hence, can be used to predict the change in stock prices for certain time horizons.
With the availability of public LOB datasets, multiple machine-learning stock price prediction models have been proposed in the literature.
These architectures range from traditional machine learning techniques like support vector machines (SVMs) to deep learning architectures like CNNs, LSTMs, ConvLSTMs and attention-based architectures~\cite{8081663, ntakaris2018benchmark, zhang2019deeplob, tran2018temporal, tsantekidis2017forecasting, tsantekidis2020using}.
The authors in the above papers utilise the generalisable nature of deep learning architectures for trend prediction with high accuracies.
However, the majority of the papers fail to report their deployment latency numbers to evaluate the performance of their models in HFT environments where the time interval between two events can be in the order of nano- and microseconds~\cite{zhang2019deeplob}.


\section{Development Stages}\label{sec:development}
In this section, we outline the different stages in LSTM development, from the front-end ONNX representation to mapping LSTM computations onto the hardware building blocks available in FINN.
\subsection{Frontend: QLSTM QCDQ representation}
ONNX is an open standard designed to represent machine learning models. 
It defines a comprehensive set of operators (like \emph{Matmul}, \emph{ReLU}), which serve as the fundamental building blocks for constructing machine and deep learning models. 
ONNX aims to provide a unified format that fosters interoperability, enabling seamless integration and interaction between various ML tools and frameworks~\cite{onnx}.
The ONNX representation also serves as a foundation for various ML hardware development frameworks, such as FINN. 
Additionally, ONNX supports modeling quantisation operators, such as \emph{QuantizeLinear} and \emph{DequantizeLinear}, enabling the representation of QNNs with standard ONNX operators.

\begin{figure}[!t]
    \includegraphics[width=\linewidth,trim=10 12 26 10,clip]{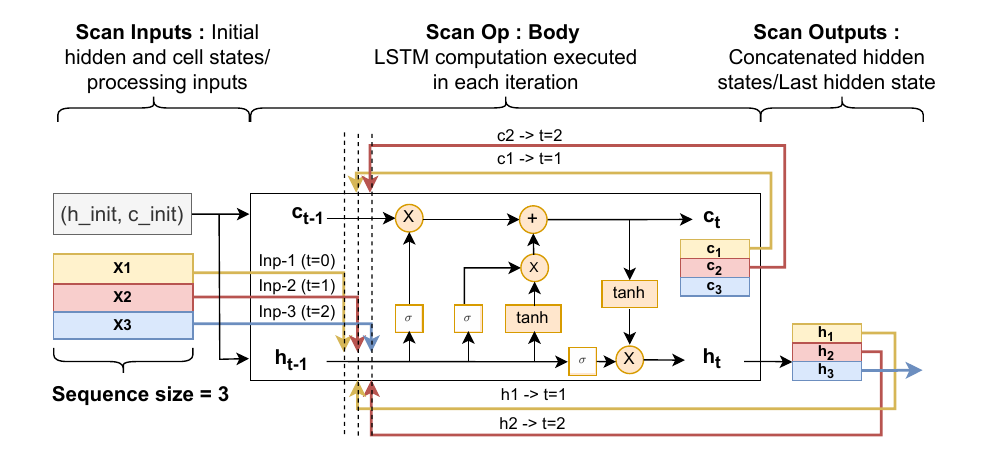}
    \caption{The figure illustrates how the \textit{Scan} operator replicates the recurrent hidden state computations in an LSTM layer. The yellow, red, and blue arrows highlight the pathways for input data processing and state updates at each time step for every input in the sequence.}
    \label{fig:scan-op}
\end{figure}

However, ONNX traditionally supports operators designed for forward computation and lacks extensive support for recurrent computations. 
ONNX recently introduced the \emph{Scan} operator to facilitate modeling of recurrent connections~\cite{scanop}.
This operator processes one or more input tensors iteratively to generate zero or more output tensors. 
It is inspired by general recurrences and functional programming constructs, such as \emph{scan}, \emph{fold}, \emph{map}, and \emph{zip}.  
During each iteration, the inputs to the operator consist of the current values of its state variables and the specific element of the input tensor being processed. 
Its outputs comprise the updated state variables and one or more output tensors.  
The computation performed in each iteration is defined by the operator's \emph{body} attribute that encodes a computational graph.
The operator has unique properties that differentiate it from other ONNX compute nodes:
\begin{itemize}
    \item It allows the state variable to be updated after each iteration, enabling the updated state to be used in the processing of the next input.  
    \item It processes inputs sequentially, either row by row or column by column, continuously updating the hidden state for each input while storing the hidden states at each step.  
\end{itemize}  
We utilise this container operator to encapsulate the LSTM compute graph within its \emph{body}, effectively modelling the recurrent computational structure of the LSTM layer. 
The quantised ONNX implementation of the layer is constructed using the \emph{QuantizeLinear}, \emph{Clip}, and \emph{DequantizeLinear} operators, forming a QCDQ representation that exactly follows the QuantLSTM implementation in \emph{Brevitas} (a framework for training quantised neural networks~\cite{brevitas}). 
To verify its functional correctness, we compare our quantised LSTM layer implementation against this QuantLSTM layer.
Figure~\ref{fig:scan-op} illustrates the internal workings of the \textit{Scan} operator in the context of the LSTM compute graph over three time steps.

The primary advantage of developing this QCDQ-based implementation using the \textit{Scan} ONNX operator is that it provides precise control over the bit-widths of all $11$ internal quantisers within the QuantLSTM layer~\cite{brevitas_quantlstm}.
This capability directly facilitates the rapid development and prototyping of mixed-precision quantised LSTM-based models, enabling fine-grained exploration of trade-offs between model accuracy and hardware efficiency. 
In contrast, the standard LSTM operator in ONNX does not expose the internal computational structure of the node, making such customisations infeasible.

\subsection{FINN: Compiler Transformations}
After functionally verifying the QCDQ LSTM layer implemented with the \textit{Scan} operator, the QCDQ computation graph should be transformed to a set of pre-built hardware building blocks from the \textit{finn-hlslib} kernel library. 
At the time of this project, the \textit{finn-hlslib} kernels were limited to supporting integer-only compute. 
However, the quantised graph contains floating-point operations coming from scale and bias operations of the quantisers. 
To achieve a functionally equivalent integer-only compute graph, we leverage FINN's  \emph{streamlining} process, which moves floating-point operations towards thresholding layers and merges them by updating the threshold values~\cite{umuroglu2017streamlined}.
To achieve this, the QCDQ graph is transformed to present quantisers as thresholding layers utilising the quantised ONNX (QONNX) format.
This allows us to reuse existing conversions and transformations available within the FINN flow and adapting them when required to support the quantised recurrent representations.
We adopted the QCDQ-to-QONNX conversion because at the time of this work Brevitas does not support direct functional QONNX export for QLSTM layers. 
Instead, we used Brevitas solely to extract weights and quantisation parameters, constructing a custom QONNX graph from the corresponding QCDQ nodes.
QONNX project extends ONNX framework by introducing custom quantisation operators (\textbf{Quant}, \textbf{BipolarQuant}, \textbf{Trunc}) to support arbitrary-precision uniform quantisation~\cite{pappalardo2022qonnx}. 
The conversion from QCDQ is achieved by applying the \texttt{ConvertQCDQtoQONNX} transformation to the graph. 
This transformation converts the QCDQ operators into \emph{Quant} nodes, as shown below:  
\begin{equation}
\scalebox{0.8}{$\texttt{QuantizeLinear + Clip + DequantizeLinear} \rightarrow \texttt{Quant}$}
\end{equation}
The QONNX LSTM graph is subsequently converted to the \emph{FINN-ONNX} representation for further processing through the FINN flow. 
%
This conversion is performed using the \texttt{ConvertQONNXtoFINNONNX} transformation, which replaces QONNX operators and activations with the \emph{matmul}, \emph{multithreshold} operators, as shown below:  
\begin{equation}
\scalebox{0.9}{$\texttt{Tanh + Quant} \rightarrow \texttt{Multithreshold} + \texttt{Mul} + \texttt{Add}$}
\end{equation}
Finally, the  LSTM computation graph also includes \emph{sigmoid} and \emph{tanh} activations, which are not directly supported in the above transformation. 
However, since they are both monotonically increasing, we model these using \emph{threshold} operators.
We developed functions that will generate thresholds for \emph{tanh} and \emph{sigmoid} activations utilised in the \texttt{ConvertQONNXtoFINNONNX} tranformation to achieve the the result shown in eq $8$.
Subsequently the graph is functionally verified and passed to downstream layers for further streamlining. 

\begin{figure}[!t]
    \includegraphics[width=\linewidth,trim=32 12 22 20,clip]{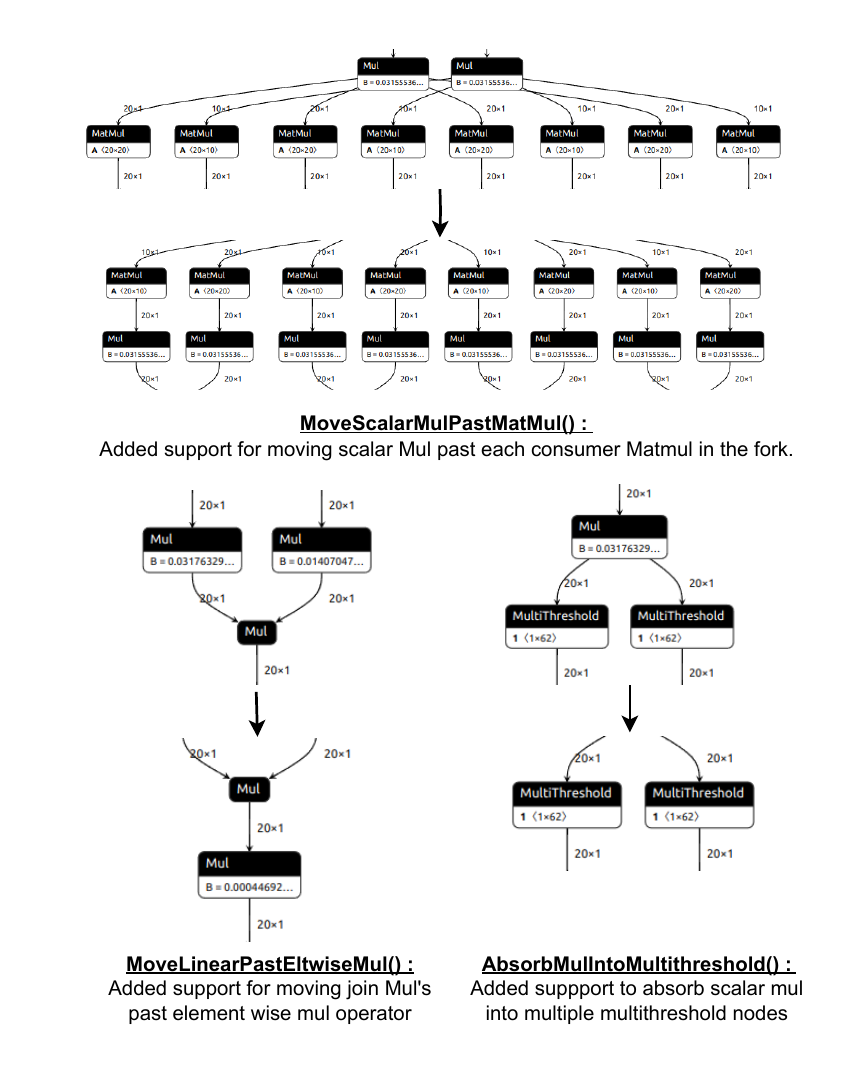}
    \caption{The figure illustrates examples of which transformations in the FINN compiler were modified and the result they achieved.}
    \label{fig:streamlining}
\end{figure}

\begin{figure}[!t]
    \includegraphics[width=\linewidth]{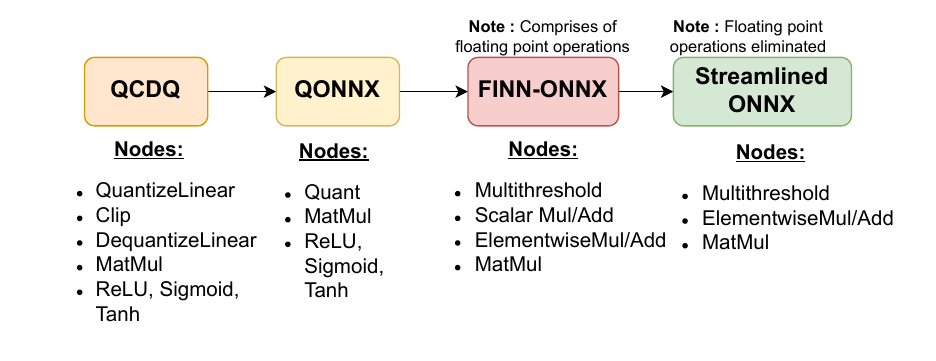}
    \caption{The figure demonstrates the compute nodes each representation comprises of in the process of obtaining the final streamlined quantised LSTM compute graph.}
    \label{fig:streamline-summary}
\end{figure}

At this stage, the primary goal is to minimise floating-point operations in the computation graph (the {\emph{Mul} and \emph{Add} nodes introduced in the previous step}) and eliminate operators by fusing/absorbing them wherever possible to reduce the resource usage of the design.  
We try to achieve this by applying pre-existing transformations to the graph wherever possible and developing transformations (new or adaptations of existing FINN transformations) for unique compute patterns that are not fully absorbed (streamlined).
Table~\ref{tab:pre_existing_transformations} describes all the transformations used in the quantised LSTM graph streamlining process. 
The transformations we developed rearrange the computations in the graph to position the floating-point operators close to the next multi-thresholding operator, allowing it to be absorbed by it.
Figure~\ref{fig:streamlining} illustrates examples where our transformations allow the floating point operations to be absorbed into the nearest multi-thresholding operator.
These transformations allow the LSTM graph to be fully streamlined, where all computations can be represented by (hardware) library functions in FINN and can hence be composed into hardware using the FINN-HLS backend. 
Figure~\ref{fig:streamline-summary} summarises the transformation process to obtain the final streamlined graph.

\begin{table}[t]
    \centering
    \caption{The table shows the list of pre-existing transformations used to transform LSTM compute graph along with new/modified transformations.}
    \begin{tabular}{ll}
        \toprule
        \textbf{Pre-existing transformations} & \textbf{New/Modified Transformations} \\
        \midrule
        MoveAddPastMul() &  MoveScalarMulPastMatMul() \\  
        MoveScalarAddPastMatMul() & MoveLinearPastEltwiseMul()\\ 
        MoveAddPastMul() &  AbsorbMulIntoMultiThreshold()\\  
        CollapseRepeatedAdd/Mul() & AbsorbsignbiasintoMultithrehsold() \\  
        AbsorbAddIntoMultiThreshold() &  \\  
        RoundAndClipThresholds() & \\
        \bottomrule
    \end{tabular}
    \label{tab:pre_existing_transformations}
\end{table}

\subsection{QLSTM HLS Backend}
The backend flow maps the QCDQ representation to a functionally equivalent graph consisting of hardware blocks that implements the operators/transformations. 
This produces a quantised LSTM (Q-LSTM) computation graph with the following operators: \texttt{Threshold}, \texttt{MatMul}, \texttt{Elementwise Add}, and \texttt{Elementwise Mul}. 
As before, we leverage pre-existing hardware blocks available in \emph{finn-hlslib} which were developed for mapping feed-forward networks (using Vitis HLS) to map the forward computation path of LSTM layers. 
In the case of recurrent (loopback) connections, we utilise a standard \emph{for} construct to map these to hardware blocks, whose bounds are determined by the input sequence length of the model.
This loop structure introduces sequential dependencies in the LSTM computation graph and imposes a limitation on parallelisation (through unrolling).
A longer input sequence length can increase the initiation interval (II) of the loop, which could be a constraint for real-time applications.
Utilising the building blocks enables us to design a generic hardware layer with customisable parameters such as data types, input size, hidden layer dimensions, and sequence length. 
This allows for simple integration in the FINN hardware generation flow and provides a flexible interface for arbitrary settings.
Once the design is complete, we synthesise the accelerator and export it as an independent IP block. 

\subsection{QLSTM layer hardware testing}
After generating the independent LSTM IP block using Vitis HLS, we integrate it with an AXI-DMA-based subsystem in the Vivado IP Integrator to build the bitstream, which is then tested on the FPGA. 
For integration and functional verification, as well as benchmarking, we follow the three-part \emph{tutorial}\footnote{\scriptsize\url{https://discuss.pynq.io/t/tutorial-using-a-hls-stream-ip-with-dma-part-1-hls-design/3344}}, which demonstrates the complete integration process.
The PYNQ APIs are used to drive the IP with input data and retrieve the output, allowing for functional verification of the hardware operation.
\section{Integration Case Study}\label{sec:experiments}
In this section, we present a case study for accelerating an LSTM-based model on an FPGA using our extensions for FINN, for predicting stock price trends using the openly available LOB dataset (normalised version). 
Our model design closely follows the architecture of the most successful approach reported in the literature for this dataset~\cite{zhang2019deeplob}. 
The details of the case study, including design choices and implementation aspects, are provided in the following subsections.

\subsection{Model development}
\subsubsection{ConvLSTM-HFT classification model}
We replicate the ConvLSTM architecture of DeepLOB~\cite{zhang2019deeplob} for our case, and evaluate the effectiveness of the proposed FINN extensions and the integration with standard FINN flow.  
DeepLOB utilises a series of convolutional blocks to extract features from a segment of order-book data, which is then passed through an LSTM layer to capture the time-series dependencies within the features. 
A final dense layer is employed for classification of stock price trend prediction.
In our case, we use two convolutional blocks to extract features from a block of 100 order-book entries, followed by an LSTM layer for time-series learning, and a dense layer at the output for classification. 
The key distinction between our implementation and DeepLOB lies in the use of inception blocks; DeepLOB incorporates inception blocks in its convolutional layers, while our model does not.
Our experiments indicate that omitting inception blocks does not result in a significant loss of accuracy, as shown in the evaluation subsection.

Each convolutional block in our model uses three Conv2D layers. 
The first block utilizes filters of sizes $64$, $32$, and $32$, with a higher number of filters in the initial layer to maximise feature extraction. 
The second block employs filters of sizes $64$, $16$, and $4$, designed to extract and preserve the most salient features for subsequent processing by the LSTM layer.
To balance effective feature extraction with computational efficiency, the first layer in each convolutional block is configured with a stride of $2$, facilitating dimensionality reduction and reducing the complexity of the subsequent LSTM layer. 
All convolutional layers utilise a $(3 \times 3)$ kernel size and are followed by a \emph{Batch Normalization} layer and a \emph{ReLU} activation function to enhance training stability.
The LSTM layer that follows the convolutional blocks is configured with $64$ hidden units, an input size of $40$, and a sequence length of $25$. 
The classification stage consists of two fully connected layers with $256$ and $3$ neurons, respectively, where the final layer corresponds to the output classes.
The complete model comprises of approximately 141K trainable parameters.
The final model configuration was derived through an iterative process aimed at balancing performance and model complexity, to achieve good model quantisation performance and resoure efficiency during the final hardware implementation. 
This exploration involved several key design decisions, including the selection of the number of convolutional blocks, the number of filters, filter sizes, and strides within each block (to optimise the input size for the LSTM layer), the number of hidden units in the LSTM layer, and the configuration of the final classification module such as the number of dense layers and the number of neurons in each layer.

\subsubsection{Quantising the ConvLSTM model}
Once the final configuration of the floating-point model was established, based on test accuracy metrics, we proceeded to quantise the model using a W8A6 (8-bit weight, 6-bit activation) scheme with the \emph{Brevitas} library~\cite{brevitas}. 
While the standard Q-Conv and Q-Linear layers in Brevitas allow quantisation of the weights, the proposed Q-LSTM layer offers additional flexibility due to its internal computation structure, enabling both weight quantisation and the selection of quantisation levels for the different activation quantisers within the layer.
In our implementation, we apply uniform quantisation scheme (W8A6) to the LSTM layer as well as the convolutional and dense layers. 
This configuration was chosen after systematically reducing the activation bit-width and observing the point at which a significant degradation in model accuracy occurred.
The next subsection provides detailed descriptions of the training process for both the floating-point and quantised models.

\subsection{Dataset and Training}
\subsubsection{Dataset}
We utilise the FI-2010 dataset, which captured LOB data for five stocks traded over a $9$-day period on the Helsinki Stock Exchange~\cite{ntakaris2018benchmark} to train the quantised-ConvLSTM model.
Each record comprises $144$ features and five labels that corresponds to future price movements over prediction horizons of $1$, $2$, $3$, $5$, and $10$ timesteps (each timestep being $\approx$ $192$ ms apart). 
The labels classify the stock's movement direction as \emph{downward}~(0), \emph{stationary}~(1), or \emph{upward}~(2) after each horizon~$k$.
The input features capture a comprehensive view of the order book, including bid/ask prices, bid/ask volumes, their derivatives, and other metrics that represent a $10$-level order book structure. 
A detailed breakdown of these features in the dataset is provided in Table~\ref{tab:feature_sets}.
Although the original dataset is not publicly available, three normalized versions — \emph{z-score}, \emph{min-max scaling}, and \emph{decimal precision} normalisation — have been made available for research purposes. 
Out of the three, we use the \emph{z-score} normalised version of the dataset which has widely adopted in the literature for predicting the trends of the stocks in the dataset~\cite{zhang2019deeplob}. 

\begin{table}[t]
    \centering
    \caption{The table describes the LOB features comprised in the FI-2010 dataset.}
    \scalebox{0.7}{
    \begin{tabular}{@{}ll@{}}
        \toprule
        \textbf{Feature Set} & \textbf{Description}  \\ \midrule
        $u_1 = \{P_{ask_i}, V_{ask_i}, P_{bid_i}, V_{bid_i}\}_{i=1}^{n}$ & 10-level LOB data \\ 
        $u_2 = \{(P_{ask_i} - P_{bid_i}), (P_{ask_i} + P_{bid_i}) / 2\}_{i=1}^{n}$ & Spread and Mid-price  \\ 
        $u_3 = \{P_{ask_n} - P_{ask_1},..., |P_{bid_{i+1}} - P_{bid_i}|\}_{i=1}^{n-1}$ & Price differences\\ 
        $u_4 = \left\{\frac{1}{n}\sum_{i=1}^{n} P_{ask_i}, \frac{1}{n}\sum_{i=1}^{n} P_{bid_i}, \frac{1}{n}\sum_{i=1}^{n} V_{ask_i}, \frac{1}{n}\sum_{i=1}^{n} V_{bid_i}\right\}$ & Price and Volume means \\ 
        $u_5 = \left\{\sum_{i=1}^{n} (P_{ask_i} - P_{bid_i}), \sum_{i=1}^{n} (V_{ask_i} - V_{bid_i})\right\}$ & Accumulated differences  \\ 
        $u_6 = \{dP_{ask_i}/dt, dP_{bid_i}/dt, dV_{ask_i}/dt, dV_{bid_i}/dt\}_{i=1}^{n}$ & Price and Volume derivation \\
        $u_7 = \{\lambda_1 \Delta t, \lambda_2 \Delta t, \lambda_3 \Delta t, \lambda_4 \Delta t, \lambda_5 \Delta t, \lambda_6 \Delta t\}$ & Average intensity per type \\ 
        $u_8 = \left\{1_{\lambda_1 \Delta t > \lambda_1 \Delta T}, 1_{\lambda_2 \Delta t > \lambda_2 \Delta T},...,1_{\lambda_6 \Delta t > \lambda_6 \Delta T}\right\}$ & Relative intensity comparison  \\ 
        $u_9 = \{d\lambda_1/dt, d\lambda_2/dt, d\lambda_3/dt, d\lambda_4/dt, d\lambda_5/dt, d\lambda_6/dt\}$ & Limit activity acceleration  \\ \bottomrule
    \end{tabular}}
    \label{tab:feature_sets}
\end{table}

\subsubsection{Training}
The z-score version of the dataset uses normalised floating-point features, which were quantised to INT8 precision for training the Q-ConvLSTM model.
From the dataset, we used the first seven days of data for training and validation purposes with $90/10$ split.
We used the remaining two days of data for testing the model, following the same training approach presented in~\cite{zhang2019deeplob}.
Quantising features into lower integer ranges ($<$ INT8) significantly hindered model convergence, as the narrow range caused many similar feature values to map to the same integer, diminishing useful information.
We started with the training of the floating-point model on the dataset to achieve a baseline model with acceptable accuracy.
We then used the trained floating-point model as a starting point to train and fine-tune the quantised version of the model, improving convergence by $10\%$ compared to training the quantised version of the model from scratch.
In all the training flows, the model with the lowest validation loss was chosen for testing.
Both the floating-point and quantised models were trained for $100$ epochs with early stopping mechanisms that triggers an exit from loop if no decrease of validation loss was observed for $20$ consecutive epochs.
We used the cross-entropy loss function and the Adam optimiser with a learning rate of $0.001$ and a batch size of $16$ for training the floating point model.
To train the quantised model, we used the same hyperparameters as the floating point version except the batch size, which was set to $32$ to reduce the training time.
Post training, we observed that the quantised model suffered from an accuracy drop of $\approx$ 1.5\%.
With further fine-tuning, where we trained this model further using a  faster ($0.01$) learning rate, the model achieved 1\% higher accuracy than the floating-point version, and was passed through to the hardware implementation phase.
Table~\ref{tab:float-quant-comp} shows the comparison of the DeepLOB, ConvLSTM and Q-ConvLSTM model trained on the FI-2010 dataset for different accuracy metrics in detail.

\subsection{Experimental Evaluation}
\subsubsection{Hardware generation and deployment}
The trained model is compiled using the proposed extensions integrated into the FINN flow to generate a dataflow accelerator IP (\texttt{mvau width} $= 36$ and \texttt{target fps} $ = 1000$) and evaluate the effectiveness of our proposed extensions and transformations. 
We chose a Zynq Ultrascale+ platform as the target (XCZU7EV device on the ZCU104 development board) mimicking an edge deployment scenario, with the tools successfully synthesizing the IP for a clock frequency of 150\,MHz. 
The IP was subsequently integrated using the IPI flow to evaluate hardware performance.
From our testing, we determine the batch-1 processing latency for the quantised ConvLSTM network to be 4.3 ms, excluding any data movement overheads, identical to the cosimulation results observed in Vitis HLS.
For the specific application of trend prediction, the average interval between consecutive events in the dataset is approximately 192 ms~\cite{zhang2019deeplob}. 
Thus, the network processes data significantly faster than the required decision-making window of ten time steps ( $\approx$ 2 seconds), providing enough time for stakeholders to analyse and respond to market changes and make informed decisions.
Higher operating performance can be achieved if the network is further unrolled, targeting a datacentre class FPGA (such as the Alveo), and by replicating the LSTM cell (reducing the sequential dependency), at the cost of higher resource consumption.  
For the Zynq deployment, the resource utilisation of our design is summarized in Table~\ref{tab:utlisation}. 
The model uses approximately 49\% of LUTs, 13\% of flip-flops (FFs), and 15\% of DSP blocks, leaving room for additional network unfolding to achieve further latency reduction, if required.
More importantly, we were able to determine that the proposed mapping flow maintains a one-to-one mapping between the trained Brevitas model and the generated hardware, demonstrating the functional effectiveness of the applied graph transformations.

\subsubsection{Accuracy}
To further assess the accuracy of the generated Q-ConvLSTM IP, we evaluate its performance using \emph{precision}, \emph{recall}, and \emph{F1-score} metrics for each of the three trend prediction labels: downward, stationary, and upward. 
We compare our quantised implementation to both the floating-point ConvLSTM model (in PyTorch) and the state-of-the-art DeepLOB model to determine the impact of quantisation on its classification performance.
Table~\ref{tab:float-quant-comp} presents this comparative analysis, illustrating that the generated quantised IP maintains an F1-score comparable to its floating-point counterpart and the DeepLOB across all three trend categories. 

Table~\ref{tab:accu-comp} summarises the comparative evaluation of Q-ConvLSTM IP generated by the proposed flow relative to existing approaches in the literature, in terms of inference accuracy. 
We replicated the DeepLOB model and independently reproduced the results for our comparison, following the same training pipeline specified in the original DeepLOB paper~\cite{zhang2019deeplob}.
We also use macro F1-score as the primary evaluation metric, as done by approaches in the literature, to mitigate the risk of inflated accuracy metrics that could result from the inherent class imbalance in the dataset.
Our results show that the generated Q-ConvLSTM IP through our FINN extensions demonstrates superior prediction performance over several widely used financial prediction models. 
Specifically, it outperforms: SVM~\cite{8081663} by 42\%, MLP~\cite{8081663} by 29\%, CNN-I~\cite{tsantekidis2017forecasting} by 23\%, LSTM~\cite{8081663} by 11\%, CNN-II~\cite{tsantekidis2020using} by 33\%, B(TABL)~\cite{tran2018temporal} by 8\%.
The Q-ConvLSTM IP achieves performance equivalent to the C(TABL)\cite{tran2018temporal} and DeepLOB\cite{zhang2019deeplob} models, both of which incorporate sequential processing layers in their architectures. 
This result is particularly noteworthy as it demonstrates that quantised LSTM models can match the performance of floating-point counterparts, and that these quantised variants can be efficiently mapped to resource constrained devices using our flow. 

\begin{table}[t]
    \centering
    \caption{Performance comparison of \emph{DeepLOB}, \emph{ConvLSTM} and \emph{Q-ConvLSTM} model for the three different labels in the FI-2010 dataset.}
    \scalebox{0.85}{
    \begin{tabular}{@{}llccc@{}}
        \toprule
        \textbf{Class} & \textbf{Metric} &   \textbf{DeepLOB (\%)}  & \textbf{ConvLSTM (\%)} & \textbf{Q-ConvLSTM (\%)} \\ \midrule
        \multirow{3}{*}{Downward-0} & Precision & 76.50 & 75.20  & \textbf{76.66} \\ 
                           & Recall   & 76.68& 77.26 &  \textbf{78.44}\\ 
                           & F1-Score & 76.59& 76.22 &  \textbf{77.54}\\ \midrule
        \multirow{3}{*}{Stationary-1} & Precision & 82.92& 81.66  & \textbf{83.25}\\ 
                           & Recall   & 77.95& 71.23 & \textbf{73.50} \\
                           & F1-Score &80.36 & 76.09 &  \textbf{78.07}\\ \midrule
        \multirow{3}{*}{Upward-2} & Precision &74.69 & 72.36 & \textbf{73.53}\\
                           & Recall   &78.77 & 79.06 &  \textbf{79.93}\\ 
                           & F1-Score & 76.68& 75.56 &  \textbf{76.60}\\ \bottomrule
    \end{tabular}}
    \label{tab:float-quant-comp}
\end{table}

\begin{table}[t]
    \centering
    \caption{Comparison of the proposed Q-ConvLSTM model against the proposed state-of-the-art HFT models in the research literature.}
    \label{tab:results_zscore}
    \begin{tabular}{lcccc}
        \toprule
        \multirow{2}{*}{Model} & Accuracy (\%) & Precision (\%) & Recall (\%) & F1 (\%) \\
        \cmidrule{2-5}
        &\multicolumn{4}{c}{Prediction Horizon $k = 10$} \\
        \midrule
        SVM~\cite{8081663}  & - & 39.62 & 44.92 & 35.88 \\
        MLP~\cite{8081663} & - & 47.81 & 60.78 & 48.27 \\
        CNN-I~\cite{tsantekidis2017forecasting} & - & 50.98 & 65.54 & 55.21 \\
        LSTM~\cite{8081663}  & - & 60.77 & 75.92 & 66.33 \\
        CNN-II~\cite{tsantekidis2020using}  & - & 56.00 & 45.00 & 44.00 \\
        T-BoF~\cite{passalis2018temporal} &  66.50  & 44.60 & 68.30 &  44.70\\
        B(TABL)~\cite{tran2018temporal}  & 78.91 & 68.04 & 71.21 & 69.20 \\
        C(TABL)~\cite{tran2018temporal}  & 84.70 & 76.95 & 78.44 & 77.63 \\
        DeepLOB~\cite{zhang2019deeplob} & 77.78 & 78.04 & 77.80 & 77.88 \\
        \textbf{Q-ConvLSTM}  & \textbf{77.37} & \textbf{77.81} & \textbf{77.29} & \textbf{77.40} \\
        \bottomrule
    \end{tabular}
    \label{tab:accu-comp}
\end{table}

\begin{table}[t]
\begin{center}
\caption{The resource consumption of the Q-ConvLSTM accelerator on the ZCU104 FPGA.} 
\scalebox{0.9}{%
\begin{tabular}{@{}lccccc@{}}
\toprule
\textbf{Q-ConvLSTM} & \textbf{LUTs} & \textbf{FFs} & \textbf{LUTRAM}  & \textbf{BRAMs} & \textbf{DSPs}\\
\midrule
\midrule
Conv Network & 65091 & 36398 & 4596 & 96  & 252  \\
LSTM Network & 49439  & 22601 & 6631 & 31.5  & 13 \\ \midrule
Overall & 101489 & 58999  & 11227  & 127.5  & 265  \\
\% & 49.6 & 12.8 & 11 & 40.8 & 15.2 \\
\bottomrule
\end{tabular}} \vspace{-5mm}
\label{tab:utlisation}
\end{center}
\end{table}

\subsection{Comparison against other frameworks}
A comparable approach for mapping LSTM models to FPGAs is available through the \emph{hls4ml} flow~\cite{khoda2023ultra}. 
While both methods offer a generalised flow for mapping an LSTM accelerator on FPGA, our proposed flow using FINN offers a few key advantages.
A key distinction is that our approach constructs a quantised ONNX representation of the LSTM compute graph, modeling its functionality using the \emph{Scan} operator. 
This provides full control over all quantisation parameters, enabling mixed-precision quantisation for the LSTM layer -- an aspect that, to our knowledge, is not available using the \emph{hls4ml} flow.

Additionally, our approach effectively leverages existing FINN compiler transformations and introduces few new ones to generate a computation graph that is entirely free of floating-point operations.
This enables primary computations within the graph to rely solely on integer arithmetic to achieve significant resource savings during the hardware generation phase, allowing seamless deployment on resource-constrained FPGAs, as demonstrated in our integration case study. 
In contrast, \emph{hls4ml} retains floating point operators in the compute graph, and has primarily targeted LSTM deployments on large FPGA fabrics. 
The use of fixed-point datatypes with fractional bits in \emph{hls4ml} is effective in certain cases at the expense of increased resource consumption. 

Also, to reduce the II, \emph{hls4ml} proposes a non-static implementation of LSTM layers, where multiple LSTM compute units are instantiated to match the input sequence length. 
This design allows computed states to be passed efficiently to subsequent stages, enabling new inputs to be processed every clock cycle and achieving an II of 1. 
However, this approach results in high resource utilisation, which may limit scalability for complex models with a large number of intermediate states.
While our approach does not consider this implementation scheme at present, we recognize its potential and see value in exploring its integration with our resource-efficient design methodology.

Finally, both frameworks share the broader goal of providing a parameterisable hardware layer to streamline LSTM deployment. 
However, while our approach prioritises optimisation for constrained FPGA devices by balancing performance with efficient resource utilisation, the \emph{hls4ml} approach prioritises latency minimisation at the expense of high resource usage.

\section{Conclusion and Future Work}\label{sec:conclusion}
In this paper, we extend the FINN framework by developing support for the generalized deployment of recurrent neural networks (using LSTMs as a working example) on FPGAs.
We create a complete end-to-end framework and introduce the \emph{Scan} operator from ONNX to enable modelling of mixed-precision quantised LSTM layers within FINN. 
Additionally, we design and modify FINN compiler transformations to optimise its compute graph and efficiently map operations to hardware blocks available in the FINN backend, while supporting arbitrary integer datatypes, dimensions and sequence lengths.
We further present a case study for stock prediction using a model architecture that uses quantised LSTM layers with convolutional and dense layers (quantised ConvLSTM model).
The quantised ConvLSTM model compiled to an accelerator IP core that can be deployed on FPGAs using our extended FINN flow.
We show that our flow can generate a resource-efficient hardware accelerator IP of the model that meets the performance requirements even when mapped on an edge-class hybrid FPGA (Zynq Ultrascale+ XCZU7EV), and offers competitive inference performance when evaluated using the FI-2010 dataset for mid-price stock prediction.
In the future, we aim to explore optimisations to reduce the latency introduced by the sequential layers in the LSTM flow, taking inspiration from the \emph{hls4ml} approach (using non-static implementations for LSTM layers), and to extend the deployment pipeline to support other RNN models such as gated recurrent units (GRUs).
\section{Acknowledgements}
I would like to thank Alessandro Pappalardo and Dr. Yaman Umuroglu for their valuable insights and the engaging brainstorming sessions throughout the course of this project.

\bibliography{references}
\bibliographystyle{ieeetr}

\end{document}